\documentclass[letterpaper]{article} 
\usepackage[]{aaai25}  

\usepackage{times}  
\usepackage{helvet}  
\usepackage{courier}  
\usepackage[hyphens]{url}  
\usepackage{graphicx} 
\usepackage{natbib}  
\usepackage{caption} 
\usepackage{color}

\usepackage{amsmath}

\usepackage{array}
\usepackage{multirow}
\usepackage{mathrsfs}

\usepackage{algorithm}
\usepackage{algorithmic}
\usepackage{amssymb}
\usepackage{newfloat}
\usepackage{listings}
\DeclareCaptionStyle{ruled}{labelfont=normalfont,labelsep=colon,strut=off} 
\floatstyle{ruled}
\newfloat{listing}{tb}{lst}{}
\floatname{listing}{Listing}
\title{Position-aware Guided Point Cloud Completion with CLIP Model}
\author{
Feng Zhou\textsuperscript{\rm 1}, 
Qi Zhang\textsuperscript{\rm 1}, 
Ju Dai\textsuperscript{\rm 2}\thanks{The corresponding author}, 
Lei Li\textsuperscript{\rm 3,4}, 
Qing Fan\textsuperscript{\rm 5}, 
Junliang Xing\textsuperscript{\rm 6}
}

\affiliations{
\textsuperscript{\rm 1}North China University of Technology, Beijing, China
\textsuperscript{\rm 2}Peng Cheng Laboratory, Shenzhen, China\\
\textsuperscript{\rm 3}University of Copenhagen, Copenhagen, Denmark
\textsuperscript{\rm 4}University of Washington, Washington, USA\\
\textsuperscript{\rm 5}Skywork AI, Beijing, China
\textsuperscript{\rm 6}Tsinghua University, Beijing, China\\
zhoufeng@ncut.edu.cn, zhangqi@mail.ncut.edu.cn, daij@pcl.ac.cn\\ lilei@di.ku.dk, qing.fan@kunlun-inc.com, jlxing@tsinghua.edu.cn}

\usepackage{bibentry}

\begin{document}

\maketitle

\begin{abstract}
Point cloud completion aims to recover partial geometric and topological shapes caused by equipment defects or limited viewpoints. Current methods either solely rely on the 3D coordinates of the point cloud to complete it or incorporate additional images with well-calibrated intrinsic parameters to guide the geometric estimation of the missing parts. Although these methods have achieved excellent performance by directly predicting the location of complete points, the extracted features lack fine-grained information regarding the location of the missing area. To address this issue, we propose a rapid and efficient method to expand an unimodal framework into a multimodal framework. This approach incorporates a position-aware module designed to enhance the spatial information of the missing parts through a weighted map learning mechanism. In addition, we establish a Point-Text-Image triplet corpus PCI-TI and MVP-TI based on the existing unimodal point cloud completion dataset and use the pre-trained vision-language model CLIP to provide richer detail information for 3D shapes, thereby enhancing performance. Extensive quantitative and qualitative experiments demonstrate that our method outperforms state-of-the-art point cloud completion methods. 
\end{abstract}

%

\section{Introduction}
Point clouds are a fundamental data structure across various domains, including autonomous driving \cite{chen2022pseudo, mao20233d,cheng2024open}, robotics \cite{christen2023learning} and others \cite{oehmcke2022deep,han2020shapecaptioner,nguyen2016field,oehmcke2024deep}. However, in real-world scenarios, challenges such as object occlusions, variability in surface material reflectivity, and limitations in sensor resolution and field of view frequently result in the acquisition of incomplete point cloud data. These deficiencies impede the effectiveness of downstream applications, underscoring the critical need for point cloud completion.
Therefore, point cloud completion is indispensable, and it has attracted more research interest in recent years. 

\begin{figure}
    \centering
    \includegraphics[width=1\columnwidth]{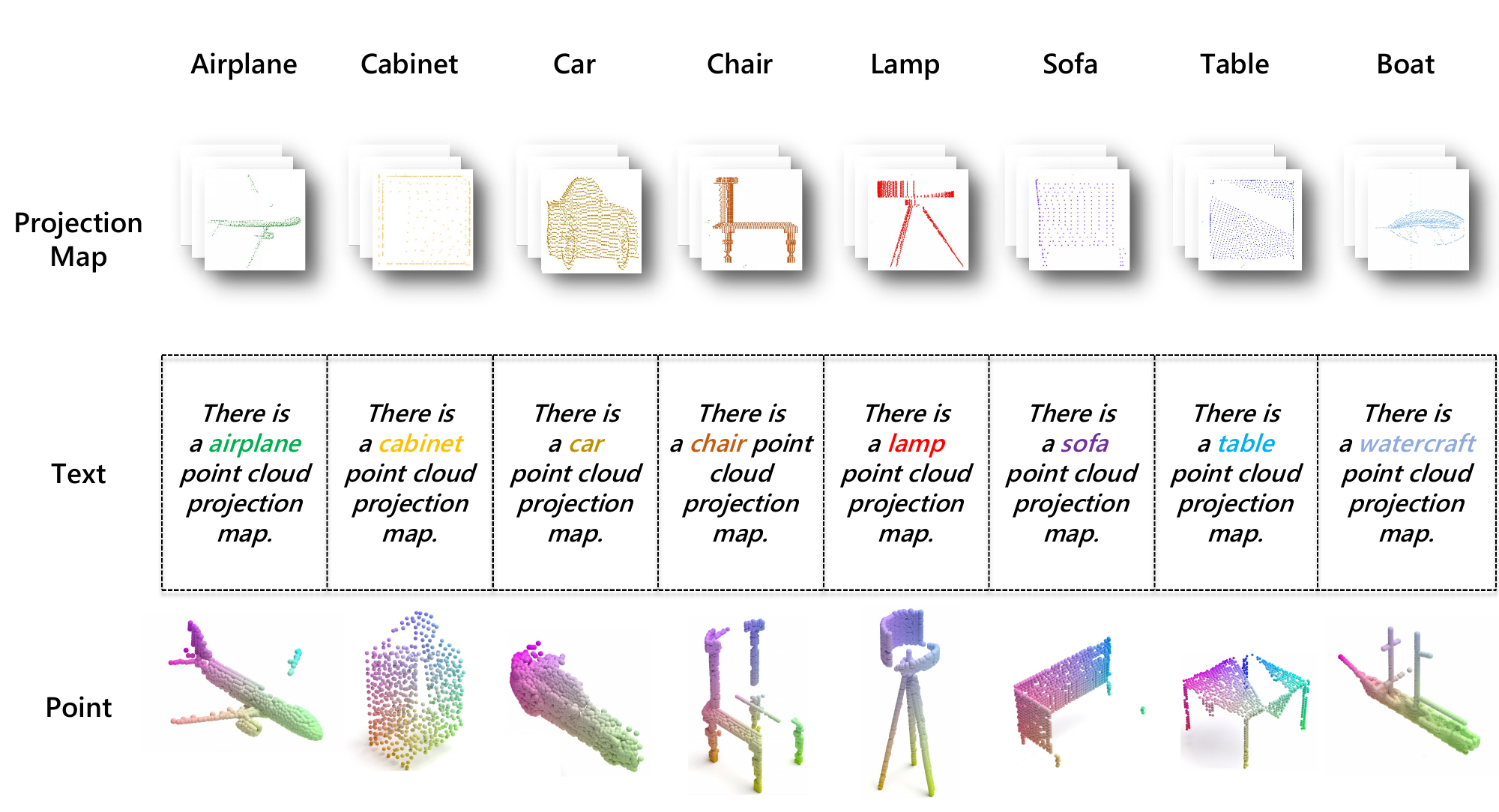}
    \caption{An exemplary instance from our PCN-TI dataset. The PCN-TI dataset comprises many triplets (Point-Text-Image) consisting of projection images, readily available textual descriptions, and incomplete point clouds.}
    \label{dataset}
\end{figure}

Leveraging large-scale point cloud datasets \cite{yuan2018pcn,yu2021pointr}, many learning-based point cloud completion methods have achieved excellent results \cite{zhang2020detail,zhang2022point,zhou2022seedformer,zhang2022shape,li2023hierarchical}. However, despite these advancements, there remains significant potential for further improvement.

The first aspect is that previous studies have identified that the missing parts either share similar structural information with the incomplete point cloud or are consistent with some existing structures. Consequently, numerous endeavors learn the prior shape information of the incomplete point cloud or suitable geometric patterns to tackle the point cloud completion challenge. For example, \cite{yuan2018pcn} proposes an encoder-decoder network to learn global shape information that directly encodes partial shapes into a global feature and then decodes it into complete shapes. \cite{tang2022lake} thinks that despite the absence of some point cloud data, the incomplete point cloud still maintains the principal skeletal structure. The authors propose a completion strategy that follows the "Keypoints-Skeleton-Shape" paradigm. By identifying and aligning key points, and then leveraging these key points with geometric priors, they introduce an innovative structure known as the surface skeleton, thereby acquiring comprehensive topological data and enhancing the information of local details.

The second aspect is that although the above methods can obtain plausible results, the incomplete point clouds may lack critical semantic parts, making it difficult for point-based networks to recognize reasonable global shapes and accurately locate missing regions. Color images can provide rich texture and color information to assist in recovering the missing geometric structure of the point cloud data \cite{zhu2023csdn}. However, pairing these images with point cloud data usually requires precise camera calibration and a complex spatial alignment process, which is very challenging in practical applications.

The third aspect is that while images can substantially aid in reconstructing the missing geometric structures of point cloud data, the unordered nature of point clouds leads to a significant challenge in precisely correlating the image information with the missing point cloud data. With the advent of visual-language pre-training (VLP) models such as CLIP \cite{radford2021learning}, ALIGN \cite{jia2021scaling} and CoCa \cite{DBLP:journals/tmlr/YuWVYSW22}, an attractive prevalence has emerged: \emph{Is it feasible to leverage VLP model to locate the missing points in point clouds?} In the field of 2D images, there are already many works have been conducted on position-guided text prompts. For example, \cite{wang2023position} introduces a method to improve the spatial understanding of VLP models by using strategically designed text prompts, which leads to better performance on vision-language tasks that require precise localization and reasoning about objects in images. \cite{zhang2021vinvl}  employ a Faster-RCNN model to extract salient region features and model the location information using bounding boxes. However, images rendered or projected from point clouds usually lack the necessary texture information, and since most images contain only a single object, it becomes challenging to accurately find the missing object parts in the image using the above techniques.

\begin{figure*}[ht]
    \centering
    \includegraphics[width=0.95\textwidth]{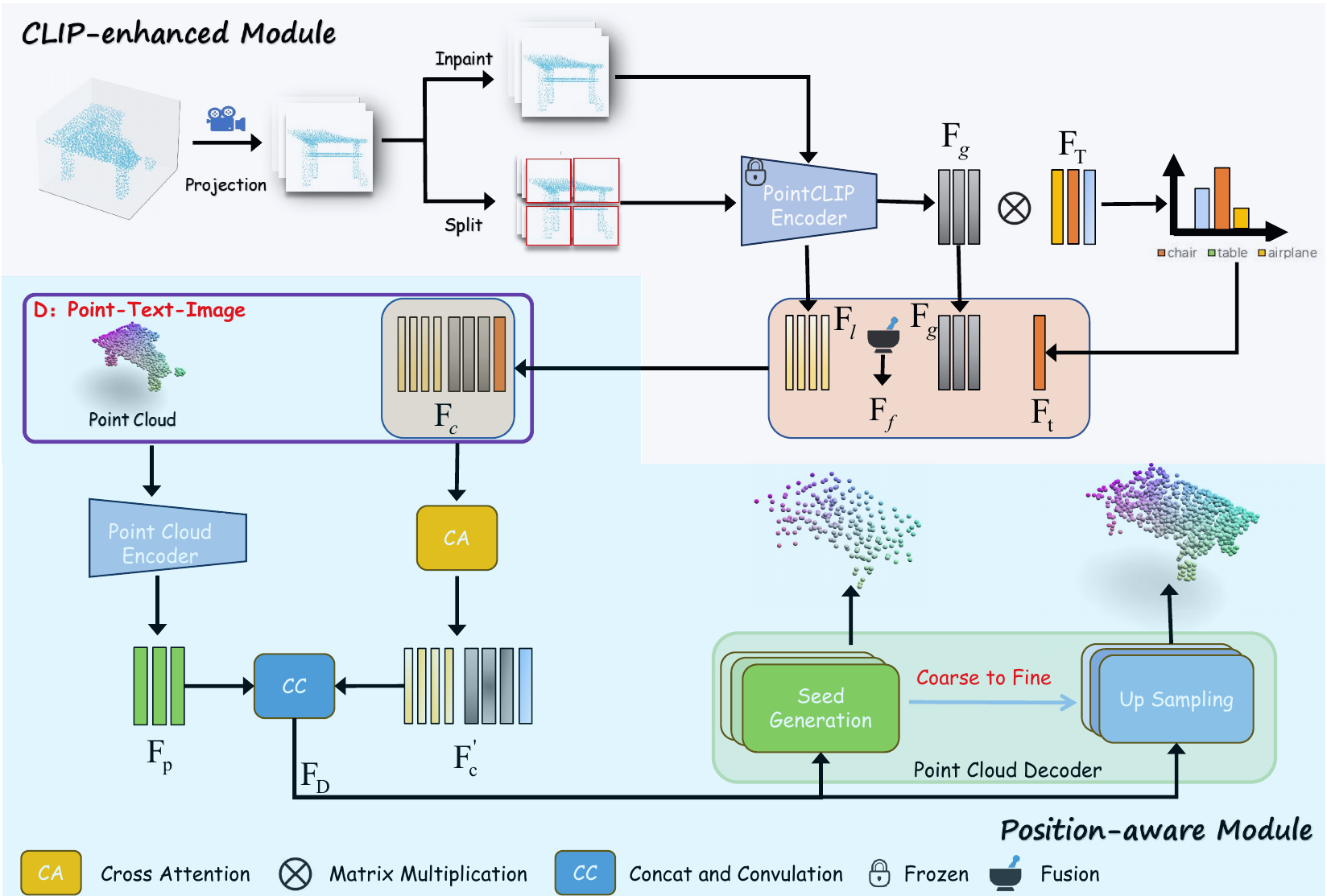}
    \caption{The overall architecture of our method consists of two main parts: the CLIP-enhanced module and the Position-aware module. $F_g$, $F_l$, $F_T$, and $F_t$ in the CLIP-enhanced module denote the global-scale feature, the local-scale feature, the text feature from CLIP, and the processed text feature of $F_T$, respectively. $F_c$, $F_c^{'}$, $F_p$, and $F_D$ denote the CLIP feature, processed CLIP feature, point cloud feature, and fusion feature fed into the decoder, respectively.}
    \label{framework}
\end{figure*}

In this paper, we propose a method that can quickly and efficiently expand an unimodal framework into a multimodal framework to address the aforementioned issues. It can simultaneously fuse the extra visual and textual information derived from the point cloud and provide precise local-global positional information for the model to ascertain where the incomplete point cloud should be completed, as shown in Figure \ref{framework}. First, unlike previous multimodal point cloud completion paradigms that focus on the fusion of point images, we introduce additional text descriptions into our model to enhance its capabilities. Drawing inspiration from \cite{song2023fine}, which introduces supplementary text descriptions for datasets comprising point-image pairs, our method differs from \cite{song2023fine} in that the textual descriptions we generate do not rely on LLMs, nor do they require additional images. Our method depends solely on the unimodal point cloud completion dataset itself, avoiding complex pre-processing steps, and is conducive to the rapid and efficient creation of a Point-Text-Image corpus, as shown in Figure \ref{dataset}. Considering that it is difficult for text descriptions to accurately locate the missing position information of incomplete point clouds, this can result in reduced performance of point cloud completion. Therefore, inspired by the work \cite{wang2023position}, which divides the image into multiple blocks, performs relevant operations on each block, and demonstrates great generalization across a variety of tasks. We also divide the projection image into different blocks and adaptively learn weight information for each block to obtain local position information to ascertain if the current block corresponds to the projection location of the missing part. Then, we inpaint the incompleted projection maps to obtain the global position information. The obtained local-global information facilitates the positioning of the incomplete point cloud. Moreover, to validate the effectiveness of the multimodal approach proposed in this paper, we conduct extensive experiments on the PCN and MVP datasets. Additionally, building upon the aforementioned method of extending from an unimodal to a multimodal framework, we introduce two multimodal datasets, PCN-TI and MVP-TI. These new datasets transform the initial unimodal datasets, which solely comprise point cloud data, into comprehensive multimodal triplet datasets encompassing point clouds, text, and images.

The main contributions can be summarized as follows:
\begin{itemize}
\item We propose a method for rapidly and efficiently transforming an unimodal point cloud completion framework into a multimodal point cloud completion framework.
\item 
We design a position-aware module to learn the location information of the missing parts of the point cloud, enabling the network to be more targeted during the completion process.
\item For each point cloud, we introduce a paired textual description and projection maps corpus, which can provide richer descriptive details, and we have proposed two extended datasets, PCN-TI and MVP-TI. Extensive experiments demonstrate the superior performance of our method against previous state-of-the-art methods.
\end{itemize}

\section{Related Work}

\subsection{Point-based Point Cloud Completion}
With the advancements in network architectures designed for point clouds, especially the emergence of PointNet/PointNet++ \cite{qi2017pointnet,qi2017pointnet++}, point-based approaches have become the mainstream solution for point cloud completion, and remarkable progress has been achieved. PCN \cite{yuan2018pcn} employs a deep neural network designed specifically for processing and completing point clouds. It adopts a coarse-to-fine architecture to generate a rough approximation of the missing parts first and then refine the details to achieve a more accurate completion. Based on the encoder-decoder architecture, many works \cite{cai2024orthogonal,wen2021pmp,wen2022pmp,xiang2021snowflakenet} obtain plausible performance. For example, SnowflakeNet \cite{xiang2021snowflakenet} interprets point cloud completion as an explicit and structured generation of local patterns and introduces a novel type of skip transformer to learn the split patterns in the Snowflake Point Decomposition (SPD). This module learns the shape context and spatial relationships between child and parent points, generating locally structured and compact point arrangements, and captures the structural features of 3D surfaces within local patches, significantly enhancing the performance of 3D shape completion. More recently, AdaPinTr \cite{10232862} and PoinTr \cite{yu2021pointr} convert the point cloud into a series of point proxies and employ a Transformer-based Encoder-Decoder for the generation of missing point clouds. Although these methods have achieved attractive results, due to the inherent limitations of singular modality, they cannot overcome the inherent ambiguity of incomplete data.

\subsection{Multimodal Point Cloud Completion}
Due to the incompleteness of the collected point clouds, the amount of information is limited, leading to significant uncertainty when inferring missing points. Additionally, point clouds are unstructured data with inherent sparsity, making it difficult to determine whether the blank 3D space is due to inherent sparsity or incompleteness, resulting in a lack of model interpretability. Many methods leverage other modalities that contain the necessary structural or semantic information of the missing parts of the given shape to complement the missing information of the point cloud, such as \cite{zhang2021view,aiello2022cross,kasten2024point,song2023fine}. ViPC \cite{zhang2021view} proposes to enhance the quality of complete shapes by utilizing the complementary modal information, integrating the local information provided by the missing point cloud with the global structural information provided by a single view to accomplish the task of point cloud completion through multi-modal fusion. \cite{zhu2023csdn} employs a coarse-to-fine completion paradigm, addressing the challenge of cross-modal data fusion through two key modules: shape fusion and dual refinement. The shape fusion module utilizes IPAdaIN to guide the geometric generation of missing areas. Subsequently, the dual refinement module enhances the accuracy of details by adjusting the positions of the generated points. 
Multimodal methods have achieved encouraging results, and our proposed method also adopts the paradigm of multimodal fusion. However, most methods expand the content of multimodal information by adding text descriptions or additional images or by combining these approaches. However, text descriptions struggle to specify the exact location of the missing parts in incomplete point clouds, and the added image requires complex alignment to be utilizable. In this paper, we adopt a self-structure-based approach on the point cloud to harness additional image information while leveraging the detailed information provided by text, combining both to obtain location information to solve this problem.


\begin{figure}[h]
    \centering
    \includegraphics[width=1\columnwidth]{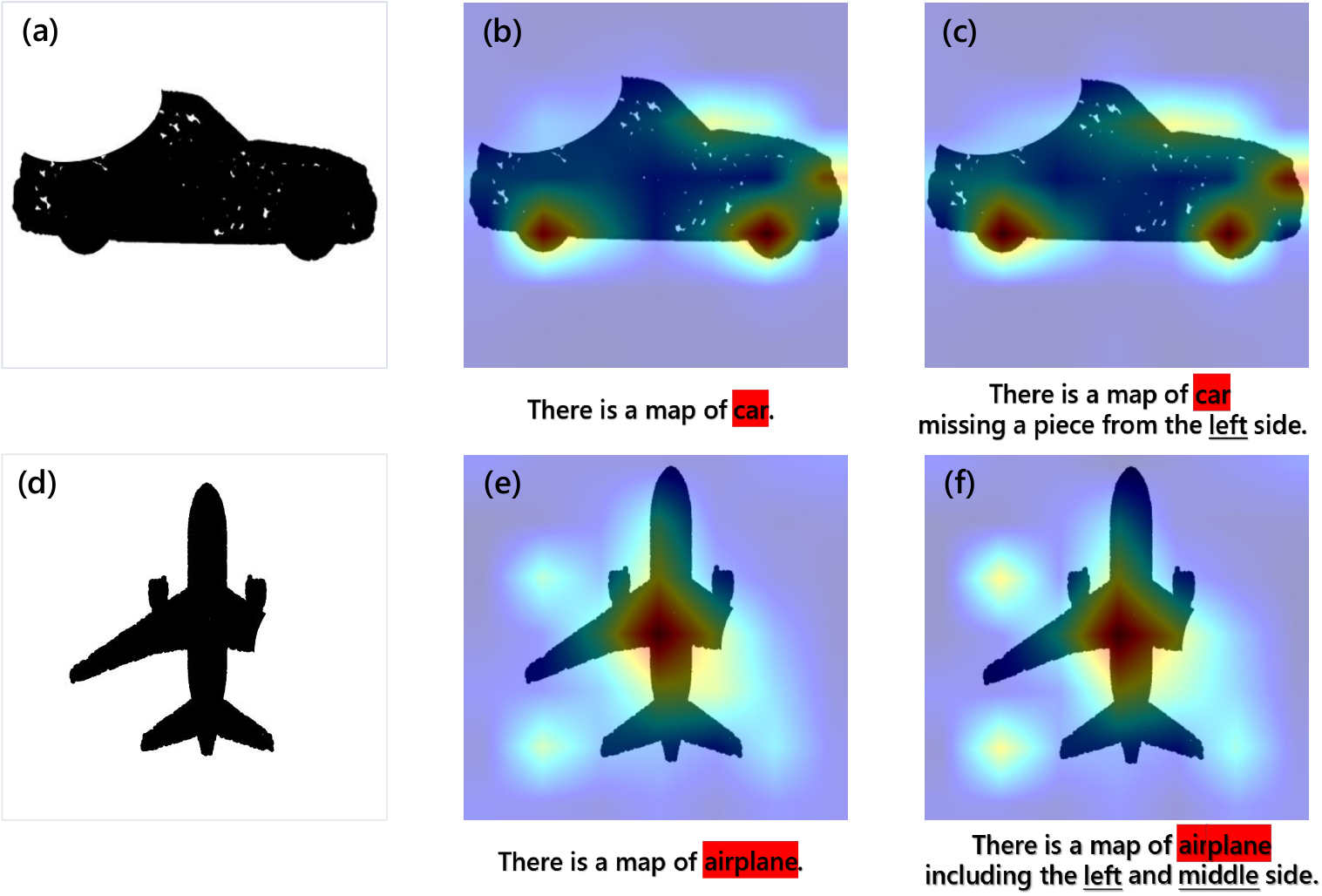}
    \caption{Our starting point for the position-aware module is as follows: (a) a projection image projected from an incomplete car point cloud, (b) the relevancy between the text "This is a map of car" and the projection map; (c) the relevancy between the text "There is a map of car missing a piece from the left side" and the projection map. (d) a projection map derived from an incomplete airplane point cloud. To illustrate the potential limitations of texts in capturing missing parts, we provide (e) and (f) for fair comparisons.}
    \label{attention-map}
\end{figure}

\section{Methodology}
\subsection{Overview}
The overall architecture of our method is shown in Figure \ref{framework}. Given an incomplete and sparse point cloud $P \in R^{N_p*3}$, the category information of $P$ is expanded to generate a sentence $T$, followed by projecting $P$ onto the coordinate axes to obtain six projection images $I = \{I_1, I_2,...,I_6\}$. The point cloud $P$, text sentence $T$, and the six projection images $I$ are then paired as a pair of data $D = \{P, T, I\}$. This preliminary preprocessing data $D$ serves as input for the network. Our goal is to infer the missing shapes of $\hat{P}$ and produce a complete and dense point cloud $O \in R^{N_o*3}$.

\subsection{CLIP-enhanced Module}
To explore the effectiveness of multimodality in point cloud completion, we provide a straightforward and practical approach to expanding existing datasets. By leveraging our proposed CLIP-enhanced module, it is possible to rapidly construct a new multimodal dataset based on current point cloud completion datasets, which can then be applied to existing CLIP methods to boost performance. Since the CLIP-enhanced module is not tailored to any specific dataset, it can be applied to common point cloud completion datasets. Here, we use the PCN dataset as an example for illustration.

We construct the text-image corpus called PCN-TI based on the PCN dataset. PCN-TI contains 30,974 pairs of text and images, forming triples along with the data in PCN, consisting of a point cloud, a text description, and a set of projection images. Some example triples are visualized in Figure \ref{dataset}. The text generation is implemented on a single NVIDIA RTX 4090. We generate the text description of each point cloud as follows: \emph {There is \{*\} point cloud projection map}, where \{*\} denotes the category of the input point cloud. Regarding the projection map, similar to \cite{zhang2022pointclip}, we perform orthogonal projections for three-dimensional coordinates (x, y, z) of the input incomplete point cloud onto six faces to obtain richer two-dimensional projection information. By assigning the depth values of each point to discrete coordinates, we form an initial projection map. To further enhance the quality of the projection map, we conduct normalization operations on each map.

Our text corpus generation in the CLIP-enhanced module is similar to \cite{song2023fine}. It enriches the expressiveness of the original dataset by incorporating textual information. Unlike \cite{song2023fine}, our method of obtaining text does not require large language models (LLMs) or complex pre-processing procedures. The text generation of \cite{song2023fine} leverages LLMs to provide very fine-grained textual descriptions. For instance, for a Lamp, \cite{song2023fine} offers a description like \emph{"This is a spiegel-symmetric lamp. The lamp has a spherical base and a circular curved shade. The lamp has one light bulb and a long stem."} The length of the generated text descriptions in their corpus ranges from 50 to 58. In contrast, our method can achieve the goal simply through keyword substitution without relying on LLMs for complex content generation. For example, the text description in our method about lamp category is: \emph{There is lamp point cloud projection map.}

Subsequently, the generated text $T$, along with the six projection images, are fed into the CLIP model. Since CLIP only accepts one text-image pair $\{T,I\}$, we utilize six identical CLIP models to process each projection image $\{I_1, I_2,...,I_6\}$ separately to get $F_t$, $F_g$, and then concatenated the obtained features $F_l$ together to form $F_c$, the detail please refer to the CLIP-enhanced module in Figure \ref{framework}.

\subsection{Position-aware Module}
While the CLIP-enhanced module provides substantial multimodal dataset support for network training, accurately identifying the missing location information in incomplete point clouds remains challenging for the network during learning. To address this issue, the initial trying is to integrate the positional information of the incomplete point cloud into the generated text description. We explore several prompts that encode positional information, for example:
\begin{itemize}
\item The projection of this \{*\} is missing a piece in the Top Right corner.
\item The top left and bottom right of the \{*\} is missing.
\item The \{*\} is missing the top left and bottom right. 
\end{itemize}
where \{*\} denotes the class, and we adopt several other similar prompts. However, we find it challenging to accurately obtain positional information on the image through text descriptions alone. To verify this hypothesis, we utilize \cite{Chefer_2021_ICCV} to visualize the attention regions of both text and images. We conduct two sets of comparative experiments. In the first, an incomplete point cloud is presented alongside a description indicating the location of the missing parts, as depicted in Figure \ref{attention-map}. By comparing Figure \ref{attention-map}(b) and (c), we observe that the positional information given in the text, "missing a piece from the left side," does not align well with the corresponding location in the image. To avoid the issue of misalignment between text descriptions and images due to the missing parts not being visible in the image, we present a comparative experiment in Figure \ref{attention-map} (e) and (f), where the text description includes positional information about existing parts in the image: "including the left and middle side." This also does not achieve a satisfactory alignment with the relevant image areas.

To effectively address this issue, inspired by the work of \cite{wang2023position}, we divide the obtained projection map into equal-size non-overlapping blocks to get good performance. The details are as follows. First, we feed the obtained projection image into the image encoder module of CLIP. The VIT-16 \cite{dosovitskiy2020vit} model is leveraged as our image encoder. The obtained projection images are segmented into $2 \times 2$ blocks in the experiments, and parameter learning is conducted for these blocks. We randomly select one block in each training iteration to learn its parameters while setting the others to a default value of 1. In testing, all well-trained parameters are loaded, and these 24 parameters are subjected to cross-attention operations with the previously acquired image features to obtain the local-scale feature $F_l$ that captures local-scale missing part information.

Additionally, in our experiments, inspired by \cite{song2023fine}, we find that although the ShapeNet-ViPC dataset does not incorporate text descriptions, it introduces complete rendering images. \cite{zhang2021view, zhu2023csdn} have already demonstrated that this additional complete rendering image provides good guidance information (such as view-guided information and image-guided information) for the point cloud, significantly enhancing the completion performance. Therefore, in our experiments, we utilize \cite{nazeri2019edgeconnect} to inpainting each projection image and then feed it into the VIT-16 to acquire a global-scale missing location feature $F_g$. 

Then, to fuse $F_l$ and $F_g$, we adopt feature fusion layers similar to \cite{zhu2023svdformer} and output the final image feature $F_f$. The local-scale feature $F_l$ is first transformed to query, key, and value tokens via linear projection and the guidance of the global-scale feature $F_g$. Then, the output feature $F_f$ is obtained after two matrix multiplication.


To this end, we feed the input point cloud $P$ into the point cloud encoder to obtain feature $F_p$. The previously obtained CLIP features $F_C$ are fed into the cross-attention module, resulting in transformed CLIP features $F_C^{'}$. We concatenate $F_P$ and $F_C^{'}$ and feed the concatenated result into a $1 \times 1$ convolutional layer to obtain fusion feature $F_D$, which serves as the input of the point cloud decoder to generate the complete point cloud.
Note that the architectures of our point cloud encoder and decoder 
depend on the baseline we choose.


\begin{figure*}
    \centering
    \includegraphics[width=1\textwidth]{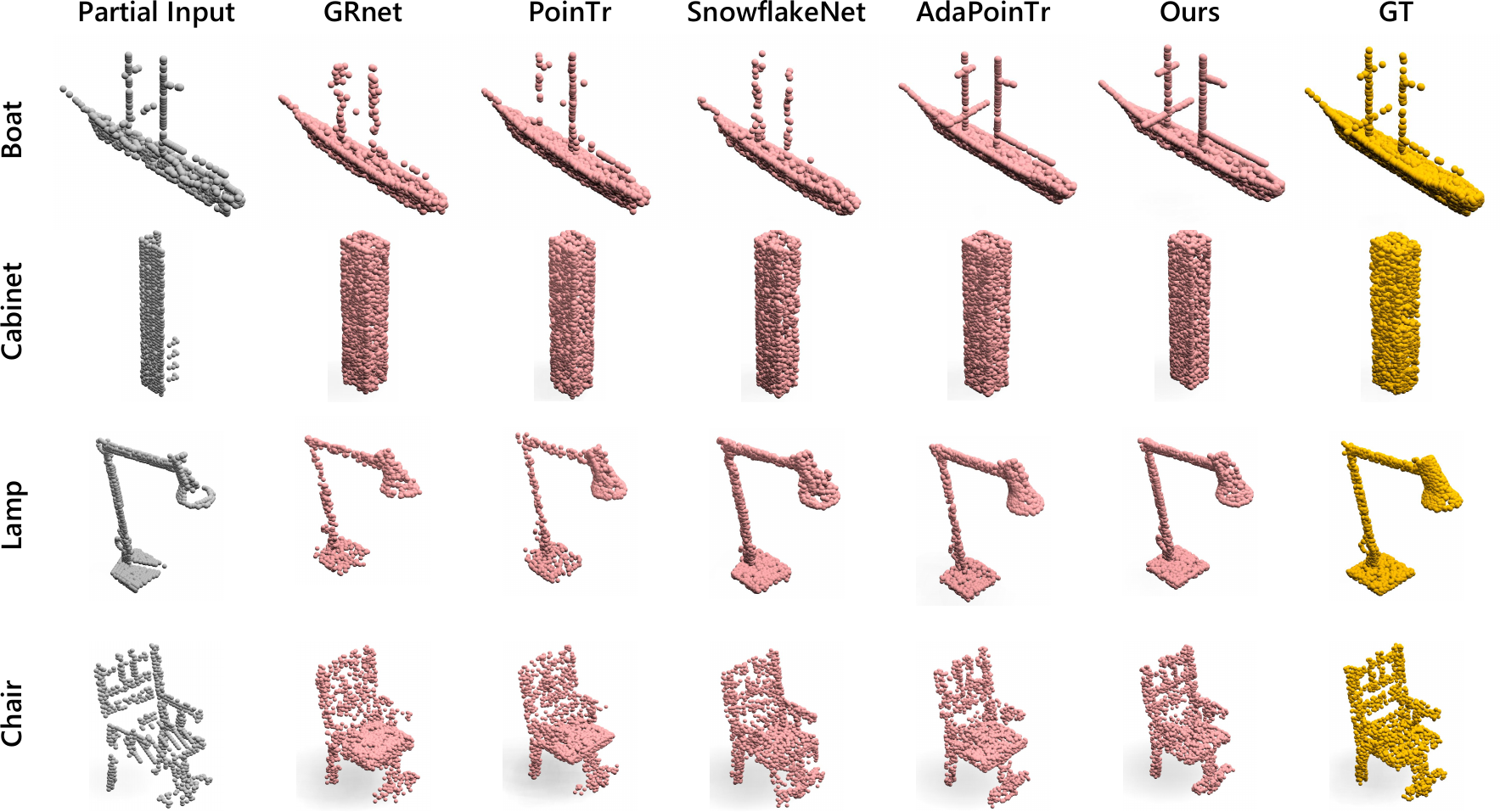}
    \caption{Point cloud completion results on PCN dataset. From left to right: partial input, results of GRNet, PoinTr, SnowflakeNet, AdaPoinTr, ours and ground truth.  Best viewed in color and zoom in.}
    \label{visualization results on PCN}
\end{figure*}

\section{Experiments}
\subsection{Datasets and Evaluation Metrics}
\textbf{PCN:} The PCN dataset \cite{yuan2018pcn} is a subset of ShapeNet dataset \cite{chang2015shapenet}, which has 8 categories and contains 30,974 pairs of partial and complete point clouds. Incomplete point clouds are generated by projecting complete shapes onto eight partial views. For each complete shape, 16,384 points are uniformly sampled from the surface of the CAD model. The dataset is partitioned similarly to PCN to ensure a fair comparison of our method with other methods. Concurrently, following prior work, the sampled points are down-sampled to a standardized size of 2,048 points for training purposes.

\begin{table}[htbp]
\centering
\renewcommand{\arraystretch}{1}
\setlength\tabcolsep{1pt}
\footnotesize
\begin{tabular}{c| c | c c c c c c c c}
\hline
Methods & Ave \(\downarrow\) & Air & Cab & Car & Cha & Lam & Sof & Tab & Boat \\
  \hline
FoldingNet  & 14.31 & 9.49 & 15.80 & 12.61 & 15.55 & 16.41 & 15.97 & 13.65 & 14.99 \\
TopNet     & 12.15 & 7.61 & 13.31 & 10.90 & 13.82 & 14.44 & 14.78 & 11.22 & 11.12\\
PCN       & 9.64  & 5.50 & 22.70 & 10.63 & 8.70  & 11.00 & 11.34 & 11.68 & 8.59 \\
GRNet      & 8.83  & 6.45 & 10.37 & 9.45  & 9.41  & 7.96  & 10.51 & 8.44  & 8.04 \\
PoinTr     & 8.38  & 4.75 & 10.47 & 8.68  & 9.39  & 7.75  & 10.93 & 7.78  & 7.29 \\
SnowflakeNet  & 7.21  & 4.29 & 9.16  & 8.08  & 7.89  & 6.07  & 9.23  & 6.55  & 6.40\\
FBNet     & 6.94  & 3.99 & 9.05  & 7.90  & 7.38  & 5.82  & 8.85  & 6.35  & 6.18 \\
ProxyFormer & 6.77  & 4.01 & 9.01  & 7.88  & 7.11  & 5.35  & 8.77  & 6.03  & 5.98\\
SeedFormer & 6.74  & 3.85 & 9.05  & 8.06  & 7.06  & 5.21  & 8.85  & 6.05  & 5.85 \\
AnchorFormer  & 6.59  & 3.70 & 8.94  & 7.57  & 7.05  & 5.21  & 8.40  & 6.03  & 5.81 \\
AdaPinTr   & 6.53 & 3.68 & 8.82  & 7.47  & 6.85 & 5.47 & 8.35 & 5.80 & 5.76 \\
ODGNet & 6.50 & 3.77 & 8.77 & 7.56 & 6.84 & 5.09 & 8.47 & 5.84 & 5.66 \\
CRA-PCN  & 6.39 & \textbf{3.59} & 8.70 & 7.50 & 6.70 & 5.06 & 8.24 & \textbf{5.72} & 5.64 \\
\hline
Ours  &  \textbf{6.34}  & 3.64 & \textbf{8.57} & \textbf{7.45} & \textbf{6.60} & \textbf{4.91} & \textbf{8.21} & \textbf{5.72} & \textbf{5.58} \\
\hline
\end{tabular}
\vspace{-2mm}
\caption{Results on PCN dataset in terms of L1 CD $\times 10^3$ (lower is better) with SOTA methods. }
\label{Performance on PCN}
\end{table}

\textbf{MVP:} The MVP dataset consists of 16 categories of high-quality pairs of partial and complete point clouds for training and testing. Each point cloud is captured from 26 uniformly distributed camera poses, offering a rich dataset for multi-view analysis and learning. Eight of the 16 categories (airplane, cabinet, car, chair, lamp, sofa, table, and watercraft) are the same as \cite{yuan2018pcn}, and another eight categories (bed, bench, bookshelf, bus, guitar, motorbike, pistol, and skateboard) are also included.

\textbf{Proposed PCN-TI and MVP-TI:} To explore the effectiveness of text descriptions and projection images in our method, we build a Point-Text-Image corpus called PCN-TI and MVP-TI based on the above two datasets. PCN-TI contains 30,974 triples, which are divided into 8 categories, the same as the PCN dataset. MVP-TI consists of high-quality pairs of partial and complete point clouds of 16 categories, the same as the MVP dataset.


\textbf{Baselines.} We evaluate four variants of pre-training models, including SnowflakeNet \cite{xiang2021snowflakenet}, PoinTr \cite{yu2021pointr}, AdapinTr \cite{10232862}, and CRA-PCN \cite{rong2024cra}, for their superior performance. If not explicitly specified, our default baseline is CRA-PCN model.

\textbf{Evaluation metrics.}
One of the challenges in point cloud completion is the comparison with the ground truth. In this paper, we follow the existing work \cite{yuan2018pcn,xiang2021snowflakenet,rong2024cra} to use the L1 CD, L2 CD, $F1$-Score@1\%, Fidelity, and MMD as the evaluation metrics.
CD (Chamfer Distance) calculates the average closest point distance between the output point cloud $O$ and the ground truth point cloud. Fidelity denotes the average distance from each point $x$ in input $P$ to its nearest neighbor in the output $O$. MMD (Minimal Matching Distance) measures how much the output resembles a typical car.

\subsection{Results on Existing Benchmarks}
\textbf{Results on PCN dataset.} We evaluate our method with many state-of-the-art (SOTA) methods, including FoldingNet \cite{yang2018foldingnet}, TopNet \cite{tchapmi2019topnet}, PCN \cite{yuan2018pcn}, GRNet \cite{xie2020grnet}, PoinTr \cite{yu2021pointr},  SnowflakeNet \cite{xiang2021snowflakenet}, FBNet \cite{yan2022fbnet}, ProxyFormer \cite{li2023proxyformer}, SeedFormer \cite{zhou2022seedformer}, AnchorFormer \cite{chen2023anchorformer}, AdaPoinTr \cite{10232862}, ODGNet \cite{cai2024orthogonal}, CRA-PCN \cite{rong2024cra}. The details are shown in Table \ref{Performance on PCN}. The quantitative results demonstrate that our method achieves the best performance across the CD metric. Specifically, our method achieves the best performance on 7/8 categories, which proves its robust generalization ability for completing shapes across different categories. Among the compared methods, PCN and SnowflakeNet are typical point cloud completion models that generate complete point clouds based on an encoder-decoder diagram via a max-pooling operation. Our method can achieve better results based on this encoder-decoder diagram. Meanwhile, PoinTr and AdaPoinTr reformulate point cloud completions as a set-to-set translation problem. Compared with these methods, our method can still perform better. Therefore, the improvements should be credited to our CLIP-enhanced module and the position-aware module, which help to overcome the limitations of traditional unimodal methods while introducing more precise information about missing positions, enabling the generation of points with greater accuracy.



Like the practice in the PCN dataset and many other works, we visually compare our method with SOTAs in Figure \ref{visualization results on PCN}. The visual results show that our method can predict the complete point clouds with much better shape quality.  
In the comparative results, it is observable that within the category of the lamp, our method generates a point distribution that is more uniform and complete, especially in the details of the lamp, such as the coil part where details are typically missing. The point cloud produced by our method is more precise in these areas. As for the chair category,  in the comparison results in Figure \ref{visualization results on PCN}, we can see that the point distribution on the chair’s back generated by our method is more uniform and complete than other methods.

\begin{table}[ht]
\renewcommand{\arraystretch}{1.25}
\setlength\tabcolsep{2.5pt}
\begin{tabular}{c | c c c c | c}
\hline
Methods & GRNet & PoinTr  & SeedFormer & SVDFormer & Ours \\
\hline
MMD \(\downarrow\) & 5.350 & 32.854 & 1.179 & 0.967 & \textbf{0.574} \\
\hline
\end{tabular}
\vspace{-2mm}
\caption{Results on LiDAR scans from KITTI dataset in terms of MMD (lower is better) metrics. The baseline is AdaPoinTr \cite{10232862}.}
\label{Performance on KITTI}

\vspace{4mm}
\centering
\renewcommand{\arraystretch}{1.25}
\setlength\tabcolsep{4pt}
\begin{tabular}{c | c c c}
\hline
Method & Points & Ave \(\downarrow\) & $F1$-Score@1\% \(\uparrow\) \\
\hline
PCN              & 2048 & 9.77 & 0.321 \\
TopNet           &2048 & 10.11 & 0.308 \\
ECG              &2048 & 7.25 & 0.434 \\
CRN              &2048 & 6.64 & 0.476 \\
VRCNet           &2048 & 5.96 & 0.499 \\
PoinTr           &2048 & 6.15 & 0.456 \\
CRA-PCN          &2048 &5.33  & 0.529 \\
\hline
Ours & 2048 & \textbf{5.32} & \textbf{0.529} \\ 
\hline
\end{tabular}
\vspace{-2mm}
\caption{Results on MVP validation set in terms of L2 CD (lower is better) and $F1$-Score@1\% (higher is better) metrics. Inputs and outputs contain 2048 points. }
\label{Performance on MVP}
\end{table}

\textbf{Results on MVP dataset.} Table \ref{Performance on MVP} shows the comparison results of our method with PCN  \cite{yuan2018pcn}, TopNet   \cite{tchapmi2019topnet}, ECG \cite{pan2020ecg}, CRN \cite{wang2020cascaded}, CRN  \cite{wang2020cascaded}, VRCNet \cite{pan2021variational}, PoinTr  \cite{yu2021pointr}, CRA-PCN \cite{rong2024cra} on MVP dataset. We can find that our method achieves plausible performance. 

\begin{figure}[ht]
    \centering
    \includegraphics[width=1\columnwidth]{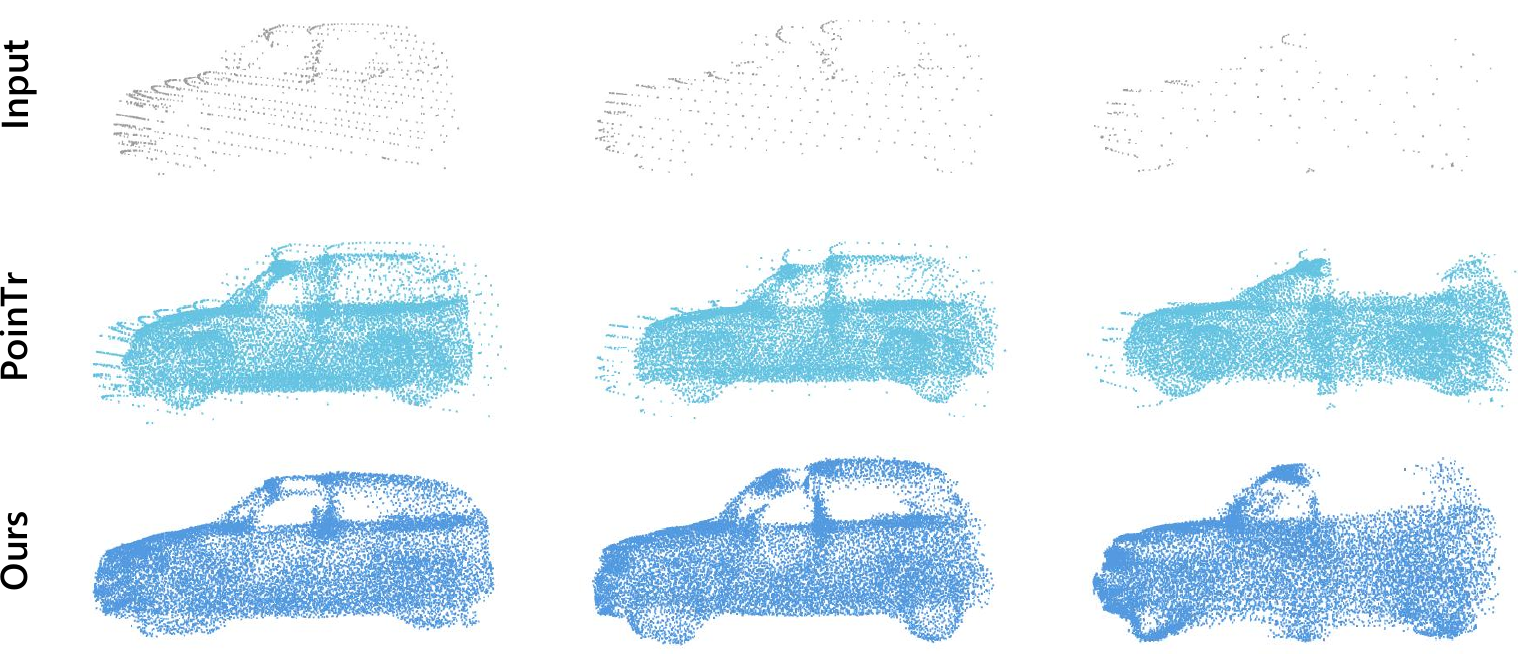}
    \caption{Qualitative results on the KITTI. From the comparison results, our method can obtain more plausible results compared with other work.}
    \label{visualization results on Kitti}
\end{figure}

\textbf{Results on KITTI dataset.} To demonstrate the effectiveness of our method in real-world scenarios, we evaluate our method on the KITTI dataset \cite{geiger2013vision} and make comparisons with SOTA methods, including, GRNet \cite{xie2020grnet}, PoinTr  \cite{yu2021pointr}, SeedFormer \cite{zhou2022seedformer}, and SVDFormer \cite{zhu2023svdformer}. We present the results using the MMD ) as a quantitative metric to assess our method without finetuning or retraining like \cite{zhu2023svdformer}, as detailed in Table \ref{Performance on KITTI}. The visual comparison in Figure \ref{visualization results on Kitti} indicates that our method also achieves superior visual results in the task of sparse point cloud completion compared to other methods.

\subsection{Ablation Study}
In this section, we implement ablation studies to demonstrate the effectiveness of the proposed CLIP-enhanced module and position-aware module. All experiments are conducted under unified settings on the PCN dataset.

\textbf{Generalizability of CLIP-enhanced module.} To validate the generalizability, we experiment with two types of text encoders: the text transformer model and the Bert-based model. We find that both models can provide robust results. 


\textbf{Effectiveness of inpainting global projection map.} To verify the impact of the inpainting global projection map on point cloud completion, we conduct corresponding ablation studies. We replace the incompleted projection map with the projection map of the complete point cloud. It indicates a significant enhancement in the point cloud completion effect. Specifically, on the CRA-PCN \cite{rong2024cra} baseline, we gain an improvement of nearly 5\%, as shown in the last column (CRA-PCN-GT) of Table \ref{abalation study on PCN}. These comparative results reveal the importance of the complete projection map for point cloud completion. By this contrast experiment, we discover that while the local information provided by the divided images and text description can bring some performance improvement, the absence of global information leads to a certain degree of performance degradation. However, the complete projection map provides more accurate positional information, which is beneficial for the model to better predict the missing parts of the point cloud and thus improves the performance of point cloud completion.

\textbf{Superiority of the proposed framework.} To further validate our framework superiority, we conduct four different baselines, SnowflakeNet \cite{xiang2021snowflakenet}, PoinTr \cite{yu2021pointr}, AdaPoinTr \cite{10232862}  and CRA-PCN \cite{rong2024cra}, with the detailed results shown in Table \ref{abalation study on PCN}. SnowflakeNet is the classic approach that encodes incomplete points into a one-dimensional feature and then decodes the complete point cloud through a coarse-to-fine method. PointTr and AdaPoinTr redefine the point cloud completion task, considering it a set-to-set translation problem. CRA-PCN represents the current SOTA results. We select these four models as our baselines and successively integrate our proposed CLIP-enhanced and Position-aware modules into these networks. The experimental results indicate that performance improvements of varying degrees are achieved across all baselines. The details are shown in Table \ref{abalation study on PCN}.

\begin{table}[t]
\centering
\renewcommand{\arraystretch}{1.25}
\setlength\tabcolsep{4pt}
\small
\begin{tabular}{c c c c}
\hline
\multicolumn{4}{c}{With / CE} \\
\hline
SnowflakeNet & PoinTr & AdaPoinTr & CRA-PCN \\
7.19 & 7.26 & 6.49 & 6.37  \\
\hline
\end{tabular} 
\hspace{2mm}
\begin{tabular} {c c c c c}
\multicolumn{5}{c}{With / CE + With / PA }  \\
\hline
SnowflakeNet & PoinTr & AdaPoinTr & CRA-PCN & CRA-PCN-GT \\
7.17 & 7.20 & 6.48 & 6.34 & 6.11 \\
 \hline
\end{tabular}
\vspace{-2mm}
\caption{Ablation study on PCN dataset in terms of L1 CD $\times 10^3$ (lower is better). 
CE and PA denote the CLIP-enhanced and Position-aware modules, respectively.}
\label{abalation study on PCN}
\end{table}


\section{Conclusion}
In this paper, we propose a method for efficiently conveying an unimodal point cloud completion framework into a multimodal point cloud completion framework based on the CLIP model. The main motivation of our work is first to introduce a general method that can generate  triple multimodal data of point cloud, textual description, and six projection images. Then, each projection map is divided into four equal non-overlapping blocks. We design a position-aware guided module that can learn a loca-global weighted map to identify the missing condition in each block. Extensive experiments demonstrate that the proposed dataset and position-aware guided module can improve our method of understanding the semantics and position information of the input point cloud. However, our method still has some drawbacks, such as the position information and the text description not being together, which will limit the performance of the model. In the future, we will further explore the relationship between the text descriptions and the incomplete point cloud, and build a fine-grained text-guided 3D point cloud completion framework.

\bibliography{aaai25}

\end{document}